# Divide and…conquer? On the limits of algorithmic approaches to syntactic-semantic structure

Diego Gabriel Krivochen

University of Reading, CINN

d.g.krivochen@pgr.reading.ac.uk

Abstract:

In computer science, *divide and conquer* (D&C) is an algorithm design paradigm based on multi-branched recursion. A D&C algorithm works by recursively and monotonically breaking down a problem into sub-problems of the same (or a related) type, until these become simple enough to be solved directly. The solutions to the sub-problems are then *combined* to give a solution to the original problem. The present work identifies D&C algorithms assumed within contemporary syntactic theory, and discusses the limits of their applicability in the realms of the syntax-semantics and syntax-morpho-phonology interfaces. We will propose that D&C algorithms, while valid for *some* processes, fall short on flexibility given a 'mixed' approach to the structure of linguistic phrase markers. Arguments in favour of a computationally mixed approach to linguistic structure will be presented as an alternative that offers advantages to uniform D&C approaches.

Keywords: syntax; natural languages; divide and conquer; phrase structure

## 1. Introduction:

Optimization problems, of the kind that have many *possible* solutions but only one *optimal* solution, are often solved in computer science by means of *dynamical programming* methods. In general, these methods combine solutions to sub-problems that are analogous to different extents to the main problem, in order to get an overall solution. Such a procedure requires, in the first place, a characterization of what *optimal* solutions look like ('characterize the structure of an optimal solution', in the words of Cormen et al., 2009: 359), as opposed to the characterization of a *single* acceptable output. Characterizing a solution for each sub-problem, plus a step that combines those characterizations, can help solve (or solve altogether) a complex problem that would otherwise have been impossible to tackle on its own. The core idea is that we need to address simpler instances of the general problem, and then combine their solutions, which can be obtained by different means. The steps to be followed in a standard D&C algorithm are the following:

a) **Divide** a problem of arbitrary complexity into multiple, simpler sub-problems, which are usually identical (e.g., for a system of *n* equations, linear algebra produces *n* $x \times y$ matrices).
b) **Conquer** each problem (i.e., solve it), either recursively or straightforwardly (depending on its complexity)
c) **Combine** the solutions to address the original problem

Examples of applied D&C include tasks like *n*-gram search, path optimization problems (see, e.g., Pereira and Lobo, 2012, for the $\mathbb{R}^2$ case), and sorting. For instance, given a random list of *n* numbers, a D&C strategy divides the array in half log(*n*) times until we get single elements that cannot be further divided, sorts those elements locally (in pairs), and then combines the partial sortings using the same method each time. Dasgupta et al. (2006: 60) offer the following pseudocode for the sorting problem:

function mergesort(*a*[1...*n*])

Input: An array of numbers $a[1...n]$
Output: A sorted version of this array
if n > 1:
  return merge(mergesort($a[1...\lfloor n/2 \rfloor]$), mergesort($a[\lfloor n/2 \rfloor + 1...n]$))
else:
  return $a$

Let us graph the D&C strategy for an unordered array:

1) Array: 5, 2, 7, 4, 6, 9, 1, 8

First, the string is divided until we get single elements (**divide**):

2) A. 5, 2, 7, 4 – 6, 9, 1, 8
   B. 5, 2 – 7, 4 – 6, 9 – 1, 8
   C. 5 – 2 – 7 – 4 – 6 – 9 – 1 – 8

Then, the algorithm allows us to regroup the elements, sorting them locally (**conquer**) and **combining** them recursively, as the output of each step is the input of the next:

3) A. 2, 5 – 4, 7 – 6, 9 – 1, 8
   B. 2, 4, 5, 7 – 1, 6, 8, 9
   C. 1, 2, 4, 5, 6, 7, 8, 9

The total time it takes is bounded by $O(n \log(n))$[1], with the worst-case running time being $\Theta(\log(n))$ (i.e., upper and lower bounded) depending on whether a *mergesort* or a *quicksort* algorithm is used.

More generally, any problem concerning optimal solutions can (but we are crucially *not* saying it *should*) be approached in a D&C manner. One such problem, we will argue here, is a fundamental question in linguistics, pertaining to syntactic theory, psycholinguistics, and computational approaches: *given a string of symbols, produce a structural description of that string*. If any notion of hierarchical dependency applies to natural language, that is, if natural languages are not purely and simply linear concatenations of symbols, then the assignment of *constituency* to a natural-language string is crucial. Consider the following example:

4) Who did you see?

Any account of the meaning of (4) must somehow capture the fact that [who] is an argument of [see], and that the temporal specification of [did] applies to [see] as well. A popular position, argued for by Noam Chomsky (1957, *et seq.*) and followers, is that there has been a *reordering* of a string generated directly by means of the successive application of rewriting rules of the form $\Sigma \rightarrow F$, which we will discuss shortly (it is also possible to apply reordering transformations to a string produced by a transformation itself, but let us keep it simple for the time being). This reordering is carried out by *transformational rules*. Here, we are concerned with *reordering* rules, defined as follows:

> *If the structural index of a transformation has n terms, $a_1$, $a_2$, $a_n$, it is a* ***reordering*** *transformation if its structural change has any $a_i$ as its $k^{th}$ term, or if $a_i$ is adjoined to its $k^{th}$ term, where $i \neq k$* (Ross, 1967: 427)

---

[1] As a reminder, $f(n) = O(g(n))$ as $n \rightarrow \infty$ if there is a constant $c > 0$ such that $f(n) \leq c\, g(n)$; that is, $f(n)$ grows *no faster* than $g(n)$. See Knuth (1976) for details.

Phrase structure rules, which are of the form Σ, F (i.e., rewrite nonterminal Σ as a –possibly unary- string F, for instance, S → NP, VP meaning 'rewrite S(entence) as Noun Phrase, Verb Phrase') generate strings like (5):

5) You did see who

Using (5) as our example, let us now assign each element an integer, such that (5) is to be rewritten in the most abstract way possible, with each X representing in this case a terminal node, thus getting the following string:

6) $X_1 – X_2 – X_3 – X_4$

(6) is a simplified version of what was known as a *structural description* or *structural analysis* in early transformational syntax (Chomsky, 1957: 111, ff.): an abstract representation of the *base position* (i.e., pre-transformational orders) of elements and their relations. Two reordering transformations, call them $T_1$ and $T_2$, changes the order to:

7) $X_4 – X_2 – X_1 – X_3$

$T_1$ is a transformation that takes the Wh- element [who] to initial position, $T_2$ is a transformation that yields subject-auxiliary inversion (see Bresnan, 1976 for formal discussion of the nature of transformational rules)[2]. Naturally, this gives us the surface order (back then called *morphophonemic structure*), but semantic interpretation is to be looked for in (5). Thus, a parser has to get to (6) departing from (7): in this way, an interpretation problem can be subsumed to a simple sorting problem. A sorting-type of approach to interpretation must be able to reconstruct the underlying (i.e., pre-transformational) order from the integers assigned to the elements, read off the morphophonemic structure. More recently, as we will see, the operations have changed their name but not their computational characteristics, with syntactic analysis still being basically a D&C process. In the next section, we will focus on the implementations of D&C routines within contemporary generative linguistics (transformational and non-transformational alike), and their limitations when it comes to capturing properties of natural languages at the syntax-semantics interface[3].

## 2. D&C in generative formal linguistics: structure building

Within formal linguistics, the problem of elaborating a structural description for a linguistic string has been addressed in a D&C fashion, with phrases or similar sub-units being identical in format to one another (e.g., the case of X-bar theory, which in current forms assumes uniform binary-branching trees; this trend started with Chomsky and Miller, 1963). The assumptions of structural uniformity, headedness, and D&C computability are common denominators across theories (HPSG, LFG, MGG, Simpler Syntax…) even if sometimes covert; these assumptions make Item-and-Arrangement syntactic theories grossly translatable into one another (see Müller, 2013 for detailed discussion; Sag, 2010 for discussion about this 'strong unification' approach to linguistic theory). Linguistic comprehension modelled as a reversed-engineered process, for instance, segments a string S of arbitrary complexity into $s_1$, $s_2$, $s_3$, …$s_n$, where all $s$ are identical in their computational and other formal properties: assuming a uniform approach to phrase structure, which is presented as an *optimal* solution to the structure building problem, units are *binarily* combined in a recursive manner to yield uniform binary branching trees (see Kayne, 1994; Chomsky, 2013, 2015, among others for developments within Mainstream Generative Grammar (MGG)).

[2] Strictly speaking, the auxiliary [do] is also inserted by a transformation, but we are assuming here it was base-generated just for the sake of the exposition.

[3] …which is *not* to say that the syntax-phonology interface is free of problems, just that we will not focus on phonological problems in the present paper.

Strict interpretations of the (Fregean) principle of compositionality within Linguistics, we propose, are derived from the underlying assumption that natural languages can be processed by means of D&C algorithms, with a string being uniformly *divided* into clauses, phrases, lexical items (the method introduced by structuralist schools in Linguistics, particularly European structuralism[4]); those units are assigned an interpretation from the level of lexical terminals up (the '*conquer*' part), and finally the solutions for the terminals (i.e., nodes which cannot be divided for some reason or another) are combined to produce a complete picture of meaning or a description of the string in linearization terms. In this sense, it is important to recall two major assumptions of MGG that Jackendoff (2011: 275) makes explicit:

- *The organization of syntactic structure is to be characterized in terms of ordered derivations that put pieces together one after another. That is, the grammar is conceived of as derivational or proof-theoretic* […].

- *Semantics is strictly locally compositional (or Fregean): the meanings of sentences are built up word by word, and the combination of word and phrase meanings is dictated by syntactic configuration.*

Jackendoff addresses the syntax-semantics relation, which is a major problem in generative linguistics. In MGG, the relation between syntax and semantics is rather a tight one: the meaning of any syntactic object SO is *fully* determined by the elements involved and the relations established between them. While the meanings of parts is a problem that has been addressed in detail, from the perspective of lexical semantics and generative semantics (see, e.g. Shibatani, 1976; Lakoff and Ross, 1973 for classic examples), the extent to which the structural template of X-bar theory has 'meaning' is far from clear. Perhaps the clearest expression of the compositionality principle in action within generative grammar is to be found in Hale and Keyser (1997)[5]:

> *We maintain that certain crucial aspects of meaning are dependent on the very structural features whose identification is at issue. If we 'knew the meaning' we would know the structure, perforce, because we know the meaning from the structure.* (Hale and Keyser 1997:40)

---

[4] In this respect, Schmerling (2016: 1) insightfully observes that

> *The French linguist Emile Benveniste (1971:101–2) speaks of a linguist's operations of 'segmentation' and 'substitution' and, further, of 'the totality of the elements thus obtained, as well as the totality of the substitutions possible for each one of them'. The latter would appear to amount to the result of what I am calling classification. (It must be noted, however, that Benveniste's European 'structuralism' was quite different from American 'structuralism' as laid out in by Zellig Harris (1951)* […]. *Benveniste goes on to discuss the impossibility of carrying out linguistic analysis without reference to meaning; he presumably intended this discussion as criticism of the stated goals of the American post-Bloomfieldian tradition.*

[5] A 1-to-1 mapping between structure and meaning (the latter, in terms of thematic relationships) was also proposed by Mark Baker (1988: 46):

> <u>*The Uniformity of Theta Assignment Hypothesis (UTAH)*</u> *Identical thematic relationships between items are represented by identical structural relationships between those items at the level of D-structure.* [i.e., before transformations apply, which relates to our previous discussion]

Mainstream generative grammar, when developing theories of lexical semantics, has remained neutral with respect to debates that are significant in philosophy, for instance, whether the principle of compositionality allows for two lexically / syntactically different sentences to be synonymous, and why different configurations in which tautologies appear can make them non-synonymous, particularly embedding (see, e.g., Carnap, 1947; examples could include tautologies under the scope of factive Vs and other opaque contexts). Rather than a truth definition, the goal of semantic studies within MGG has been identified with the representation of meaning within an ideal speaker/listener's mind, following the Competence-Performance distinction introduced in Chomsky (1965) and the further qualification that only I-language (internal, intensional, individual) is worth studying (Chomsky, 1986a). Considering that the relations allowed by the restricted generative engine in the Minimalist Program (Chomsky, 1995 and much subsequent work, particularly Chomsky, 2013) have been reduced from Head-Specifier and Head-Complement (with both heads and complements being maximal projections) to just Head-XP, Chomsky (2009: 52) has stated that:

> '*The crucial fact about Merge – the "almost true generalization" about Merge for language is that* ***it is a head plus an XP. That is virtually everything.*** (…) *For one thing it follows from theta-theory. It is a property of semantic roles that they are kind of localized in particular kinds of heads, so that means when you are assigning semantic roles, you are typically putting together a head and something. It is also implicit in the cartographical approach. So when you add functional structures, there is only one way to do it, and that is to take a head and something else, so almost everything is head-XP*.' (Highlighting ours).

Notice that the restriction on Merge follows from theory-internal considerations (theta-theory, cartography of functional material, not to mention the more basic but equally problematic notion of 'head'), and not from requirements of empirical data (a point also argued for by Müller, 2013: 921-922; as well as Jackendoff, 2008). Computationally, this restriction on the form of phrase markers (headed, binarily-branched, projecting) follows from the axiom (as it has never been empirically proven) that natural language grammars are uniformly Chomsky-Greibach normal (e.g., Chomsky, 1955, 1957), which yields production rules of the form:

> <u>Chomsky-normal:</u> every context-free language is generated by a grammar for which all productions are of the form A → BC or A → *b*. (A, B, C, nonterminals, *b* a terminal)
>
> <u>Greibach-normal:</u> every context-free language is generated by a grammar for which all productions are of the form A → *b*α, where *b* is a terminal and α is a string of nonterminal variables.

As Hopcroft and Ullman (1969: 46) claim, the original definition does not allow the empty symbol ε to appear in any Context Free Language. However, it is possible –they claim- to include the case where A → ε is a production rule (a specific case of A → *b*, for *b* = ε). If this production rule is accepted (and see Hopcroft and Ullman, 1969: 62-63 for discussion), we can briefly show both formalisms to be weakly equivalent, if A → BC can include a terminal. Trivially, A → BC is logically equivalent to A → BCε, which is in turn equivalent to A → εα, since ε is an empty terminal, and any number of nonterminals is captured as α. A more formal proof that any G-normal is equivalent to a C-normal can be derived from the results in Greibach (1965: 49) -but we will attempt no such demonstration here.

The PF (phonetic form) component of the grammar is in charge of, among other things, linearizing phrase markers for the purposes of externalization, basically, creating *strings* out of hierarchical tree graphs[6]. In this respect, Kayne (1994) offers what is now a standard view in MGG:

> '*phrase structure* (…) *always completely determines linear order* […]' (Kayne, 1994: 3)

> '*Linear Correspondence Axiom:* [LCA henceforth] *d(A) is a linear ordering of T.*' [A a set of non-terminals, T a set of terminals, *d* a terminal-to-nonterminal relation] (Kayne, 1994: 6)

*d* is a syntactic relation known as asymmetric *c-command*, applying to the set of nonterminals. In simple terms, the LCA states, on the weak interpretation, that '*when x* [asymmetrically] *c-commands y, x precedes y*' (Uriagereka, 2012: 56), and on the strong interpretations, that *x* precedes *y* ***iff*** *x* [asymmetrically] c-commands *y*, c-command being both *necessary* and *sufficient*. We will assume the strong interpretation, in the line of Kayne (1994), Moro (2000), and much related work. Following Reinhart (1976) and much related work, let us define c-command so that A *c-commands* B *iff* the first branching node that dominates A also dominates B, and neither A nor B dominates the other. *C-command* is said to be *asymmetric iff* A *c-commands* B but B does **not** *c-command* A: this qualification excludes sister nodes, which *c-command* each other.

LCA-compatible objects, under the strong interpretation (licensed by the '*…always completely determines…*' part of the LCA) are *always* and *only* of the following form:

8)

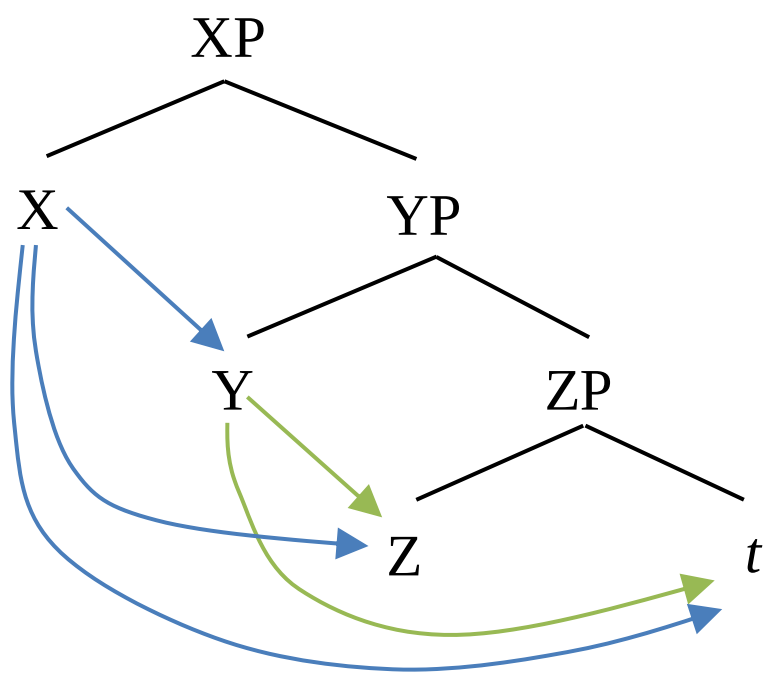

X's asymmetrical c-command path (i.e., all the nodes X c-commands but that do not c-command X) is marked in blue and Y's, in green. Z does not asymmetrically c-command anything, nor does its sister (they c-command each other, a situation which is referred to as a 'point of symmetry'), but since the most deeply embedded element in the structure is a trace *t* (which corresponds to a moved constituent), it need not be linearized (as it receives no phonological exponent), and the potentially problematic point of symmetry situation is straightforwardly avoided. Why should a point of symmetry be problematic? Because if *x* c-commands *y* and *y* c-commands *x*, there is no unambiguous way of linearizing *x* with respect to *y* while following the LCA.

The transitive c-command relations we have graphed are translated into linear order as follows:

9) X⁀Y⁀Z[⁀*t*]

[6] A reviewer has correctly pointed out that, *in principle* (this is a crucial qualification), PSR generate sets, not graphs. However, the set-theoretical approach in and of itself does not tackle the issue of linearization, which is at the very core of the reworking of X-bar theory under LCA assumptions.

The combination of the assumptions about phrase structure and linearization MGG works with can be made explicit in the following example. Assume the string (10), and the following problem: *what is the phrase marker corresponding to (10)?*

10) *abbabbababbab*

(10) is of course not any odd string we have chosen randomly, but we will come back to that below. For the time being, just assume #*a*# and #*b*#, i.e., *a* and *b* are terminals. How would an MGG algorithm analyse that? Well, let us first divide the problem into subproblems we can deal with straightforwardly. If we are only working with terminals, then we are forced by the LCA to divide the string into sets of two terminals each, *and* to add a terminal in the end of the string, which corresponds to an empty category, let us say, the null element ε. This first step, *divide*, gives us:

11) [*ab*][*ba*][*bb*][*ab*][*ab*][*ba*][*b*ε]

Then, we try to infer the syntactic structure. Let us proceed from left to right, considering the bracketed units. Given that *a* precedes *b*, according to the LCA (particularly, but not exclusively, in its strong interpretation), *a* must *c-command b*, thus, the minimal structure for the most embedded substring in (10) must be as follows for asymmetric *c-command* to hold (see Stabler, 2011 for a computational perspective on Minimalist grammars, and for notational issues as well):

12)

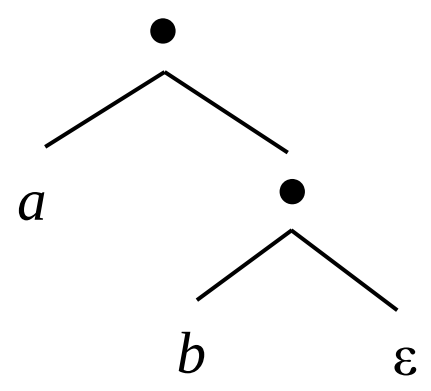

Notice that the minimal structure for an MGG phrase marker to be linearized is computationally procrustean: the assumption that the string under consideration has been generated via a Chomsky-normal / Greibach-normal grammar (Greibach, 1965) is *necessary*. Otherwise, all we get are points of symmetry, and that disrupts the LCA order. Thus, we had to add a nonterminal[7] (denoted by ●), so that there is a branching node that dominates *b* but does not dominate *a*. Moreover, since unary branching is not an option in antisymmetry-based MGG (although it is in some variants of the theory, see e.g. Adger, 2012 for a Unary Merge proposal; there are also independent arguments against unary proposals, see Krivochen, 2011, 2015a), and also because of the *trace condition* introduced above, we needed to add a terminal ε which receives no morphophonological exponent.

The previous paragraph described the second step of the algorithm, *conquer*: once the string has been appropriately subdivided, we solve the structure assignment problem for each substring, which gives us (8) for all substrings (marked with square brackets in (7)), with ε being replaced by ● (say, via a Generalized Transformation (GT); Chomsky, 1955[8]) at every step but the most embedded, as we will

---

[7] Greibach (1965: 43) refers to these as *intermediate symbols*, but this is just a terminological difference with no impact on the properties of the formal language.

[8] This might be a technicality, but let us define the steps involved in a GT, following Kitahara (1994: 49); Chomsky (1995: 189) –in turn based on Chomsky, 1955- :

i. Target a category α
ii. Add an external Ø to α, thus yielding {Ø, α}
iii. Select category β

see below in (13). The *combination* of all partial solutions for the structure assignment problem gives us the following tree graph, from which we can reconstruct (10) if we apply the LCA to linearize terminals:

13)

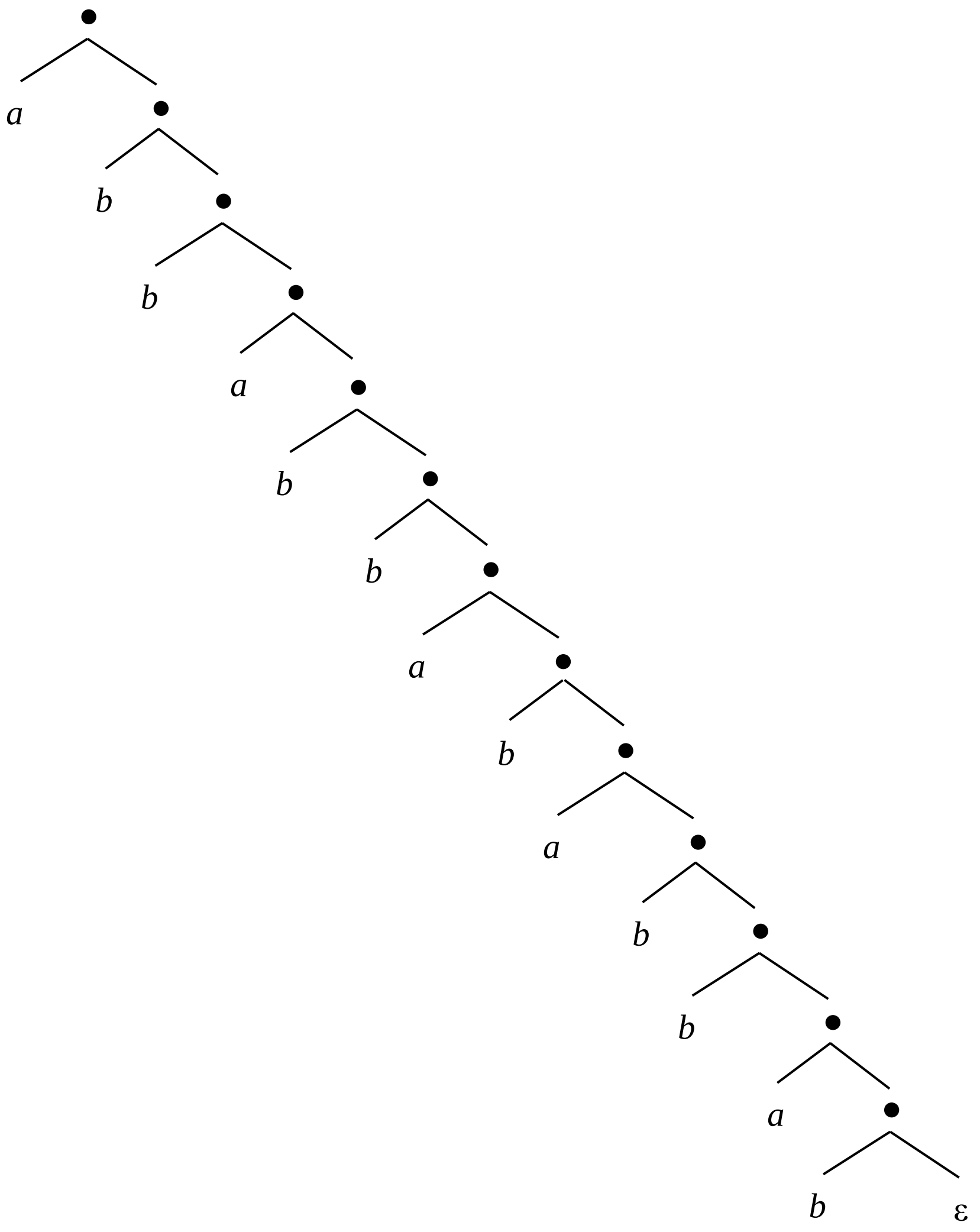

A D&C approach to the structure assignment problem forces a phrase structure grammar into a string description. It is important to notice that other, non-transformational approaches like HPSG and LFG (see Green, 2011 and Nordlinger and Bresnan, 2011 respectively, for overviews), which are Unification-based, also work with uniform phrase structure grammars (Type 1 and Type 2 grammars), as well as Tree Adjoining Grammars (Joshi, 1985); even though all these alternative generative frameworks tend to be less restrictive, computation-wise, than standard X-bar accounts (see, e.g., Green, 2011: 25). For the specific case of LFG, Kaplan and Bresnan (1982) situate the generative power of an LFG between a Type 2 and a Type 1 grammar (Context-Sensitive and Context-Free languages respectively). However, if this were true, then at least some LFG strings *should* be 'pumpable', in the following sense:

---

iv. Substitute Ø by β, creating {γ, {α, β}}

In more usual terms, Merge is a GT.

*If a language L is context-free, then there exists some integer $p \geq 1$ (called a "pumping length") such that every string s in L that is longer than or equal to p symbols (i.e. with $|s| \geq p$) can be written as*

*$s = uvwxy$*

*with substrings u, v, w, x and y, such that*

*1. $|vwx| \leq p$,*

*2. $|vx| \geq 1$, and*

*3. $uv^nwx^ny$ is in L for all $n \geq 0$.*

For a developed proof, see Martin (2010: 206-208).

However, it is not at all clear that the 'pumping lemma' (PL) for CFL applies to natural languages; thus, it is not clear that natural languages (all and only natural language strings) are generable by LFGs (see also Berwick, 1984 for a discussion of language complexity that assumes that natural languages fall within Type 2 languages; also Ristad and Berwick, 1989 for a discussion of *agreement grammars* as CFGs). One of the reasons why this appears to be so, we will argue, is that there are natural-language strings which are orthogonal to the Chomsky Hierarchy, displaying characteristics of non-normal grammars (e.g., L-grammars), a kind of formalism that is particularly important in experimental Artificial Grammar Learning studies (e.g., Saddy, 2009) and to which humans seem to be particularly sensitive. Because there is no terminal-nonterminal distinction in L-systems (every member of the alphabet is *both*, depending on the side of the rewriting rule they appear in, as we will see in (14) below), non-normal grammars are likely orthogonal to the Chomsky Hierarchy –which certainly applies to normal grammars, though-, and the pumping lemma (in any of its versions) does not apply to them: there is no way of guaranteeing that the illegal substrings **aa* and **bbb* will not appear if a string generated by the L-grammar in (14) below is pumped. If some natural language substrings are non-normal, and some processes in natural languages follow an L-grammatical logic rather than a Chomsky-normal / Greibach-normal logic (for instance, the identification of syllable boundaries, see Uriagereka, 2012: 297, ff.) the non-applicability of the PL would follow: this is a problem for theories that argue that natural languages are *uniformly* CF, CS, or 'mildly CS'. These considerations aside, though, there are more problems.

Firstly, a D&C algorithm imposes a *uniform* phrase structural description on *every* string, such that the minimal nontrivial *divide* step yields a binary substring, always of the kind [H, XP], where H is a Head (a terminal) and XP is a phrase of category X, thus, a nonterminal node (Chomsky, 2009, 2013). This implies that structural descriptions for natural language strings, if obtained via D&C, will be *uniform*: every sub-problem to be solved (in this case, the relative ordering of *n* elements in the shortest possible nontrivial substring, $n > 1$) will consist of no more and no less than 2 elements, those being a [H, XP] unit, which is 'virtually everything' (Chomsky, 2009: 52). Secondly, and in relation to the previous point, it is not clear whether D&C could handle non-normal grammars, and capture their emergent properties. It is in this sense that (10) is a special string: it is the 7th generation of the following grammar:

14) $a \rightarrow b$
$b \rightarrow a, b$
Axiom: *a*

This is a very special grammar, for it yields the Fibonacci sequence as an emergent property of counting the number of terminals/nonterminals, *a*, or *b* at any given generation (for linguistically-oriented discussion, see Uriagereka, 1998: 192-193; Uriagereka, 2012; Carnie and Medeiros, 2005; Medeiros, 2008, among others). And it is relevant for our purposes because, as Saddy (2009), Shirley (2014), among others have shown, it seems we humans are exceptionally good at Artificial Grammar Learning tasks involving stimuli generated by the grammar in (14) *versus* other artificial grammars (X-OR, for instance), including oddball experiments in visual and auditory tasks. The pattern generated by (14) also appears in the X-bar template, and in metric combinatorics and metrical feet (Idsardi and Uriagereka, 2009; Uriagereka, 2012): whatever underlies the strings presented to participants, it seems there is a particular sensitivity to strings generated by means of (14), which makes it cognitively relevant as a formal grammar. If there is a 'faculty of language', either as a module or as an emergent of interaction between more general systems, it must include aspects of non-normal grammar recognition or assignment, for otherwise aspects like the sensitivity to Fib-patterns would remain unaccounted for.

Anticipating discussion in Krivochen (in preparation), and following Rozenberg and Salomaa (1980), we say that (14) is an example of a *simultaneous, non-normal rewriting system*. (14) instantiates a Lindenmayer system, 'L-system' henceforth. These systems are also of the form $\Sigma \rightarrow F$, but they have some special properties, both of which will be reviewed below:

a) Terminals / nonterminals are defined *contextually* within a rule
b) All possible rules involving elements in a representation apply *simultaneously*

Apart from those unique properties, L-systems share many of the aspects that characterize other formal grammars: they have an alphabet, a set of accepting states, and a transition function from one state to another. The peculiarities of L-grammars pertain to how they operate with those elements, which they have in common with other formal systems. In Krivochen and Matlach (2015) we have defended the idea that L-systems are orthogonal to the Chomsky Hierarchy (and thus configure a space of grammars of their own), primarily because, as Prusinkiewicz and Lindenmayer (1990) point out,

> *In Chomsky grammars productions are applied sequentially, whereas in L-systems they are applied in parallel and simultaneously replace all letters in a given word. This difference reflects the biological motivation of L-systems. Productions are intended to capture cell divisions in multicellular organisms, where many divisions may occur at the same time.*

This means that a D&C parsing forces a certain derivational history on a string that does not correspond to the *simplest* possible one that captures the string's emergent properties (the Fib sequence, in this case), because D&C algorithms are limited to normal grammars. It should be crucial to point out that, if we took any string and tried to, say, sort substrings as in (6) and (7) above, we would be assuming that the relevant string had been generated by a normal grammar. This is an assumption that is not without consequences. The applicability of D&C routines must, we argue, be relativized with respect to what is usually assumed.
In the following sections we will discuss in turn the problems of mixed phrase markers and grammar *assignment* as the core process in parsing. We do not completely reject D&C algorithms, but will limit their applicability, while arguing for a more dynamic and semantics-sensitive approach to structure assignment for natural language strings.

## 3. <u>On the (non-) uniformity of phrase structure in natural language</u>

The argument in this section will be laid out as follows: first, we will consider some empirical predictions of a D&C approach to phrase structure. Then, we will show their inadequacy, and argue in favour of a mixed model of phrase structure which is not *globally* and *uniformly* D&C parsable.

Let us briefly consider the case of coordination (which is analyzed *in extenso* in Krivochen and Schmerling, 2015). Several analyses within the Minimalist incarnation of generative linguistics (Johannssen, 1993; Progovac, 1999; Zhang, 2010; Chomsky, 2013, to name but a few), based on axiomatic considerations about headedness and binary branching, assume the following phrase marker for structures of type [A and B][9]:

15)

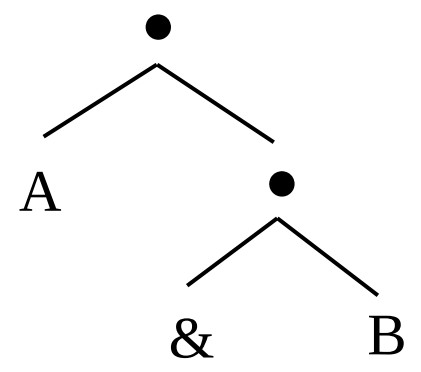

That is, a uniform phrase structural analysis is the norm for coordinated structures, and crucially, no semantic considerations affect the shape of the phrase marker. It should be clear that such a phrase marker is fully D&C compatible, for any string that contains [A and B] as a substring will be ultimately divided in [α, [β, …[A [& B]]…]] (where the identity of α and β need not concern us), following the template of (8). (15) can be LCA-linearized in a straightforward manner: A asymmetrically c-commands &, thus, A precedes &. In turn, if B is a nonterminal, then & asymmetrically c-commands the head of B. If B is a terminal, then it must either be a trace itself or be moved and leave a trace behind (since traces are not subjected to the LCA). The question is thus: does (14) correctly capture the syntax and semantics of *all* coordinated structures? Consider the following Latin examples (from Krivochen & Schmerling, 2015: 1):

16) a. Perdiderint cum me duo crimina, carmen et error (Ov. Tristia II, 207)
ruin$_{3PlPastPerf}$ with me two crime$_{Pl}$, poem and error
'Two crimes ruined me, a poem and an error'

b. effodiuntur opes, inritamenta malorum.
arise$_{3PlPastPerfImpers}$ wealth$_{Pl}$, incitements of-the-bad
iamque nocens ferrum ferroque nocentius aurum
and now harmful iron and iron$_{Dat}$ more-harmful gold
prodierat […] (Ov. Met. I, 140-142)
come-forth$_{3SgPastPerf}$

---

[9] Progovac (1999) is not concerned with linearization issues. Therefore, the point of symmetry between [&] and [B] (if [B] is a terminal) is not problematic. Chomsky (2013: 46), on the other hand, assumes there is a movement operation raising one of the conjuncts:

Input: [& [A, B]]
Output: [A [& [*t*, B]]]

In this way, the symmetry point [A, B] is eliminated. Chomsky arrives at this 'conclusion' from considerations of labelling and projection, but the final phrase marker is LCA compatible all the same. Other possibilities include assigning more structure to B, such that B = XP, and that XP branches unarily (the *Nonbranching Projection* operation in Kitahara, 1994, for instance), thus extending the tree further down and dissolving the point of symmetry.

‘There arose wealth, incitement of bad things. And now came forth harmful iron, and gold, (which is) more harmful than iron’

What we argued in that paper is that we are in the presence of two kinds of phrase markers whose computational properties differ. On the one hand, (16a) features two NPs coordinated by [et], the coordination being an epexegesis of [duo crimina], which agrees in number with the V [perdiderint]. Plural agreement, we argued, requires inner probing into the coordinated construction, which is to be taken as a nonterminal. Since we need to access the internal properties of the construction in order to account for phenomena involving the conjuncts (agreement and extraction possibilities, among others), we proposed in Krivochen & Schmerling (2015), as well as Krivochen (2015a) that the dependency between conjuncts in these cases is indeed phrase-structural, and used the term *infective* to refer to the partially transparent nature of the coordination, such that operations can target one or the other of the conjuncts, or both. (12) –or any other normal structure-, thus, seems to be a valid template for this particular kind of coordination.

On the other hand, (16b) presents a [N Nque] pattern that triggers *singular* agreement in [prodierat] (see fn. 6). Notice that the inner complexity of the NPs involved is greater than in (11a), for we have predications within those NPs: [nocens ferrum] and [ferro nocentius aurum]. The [que] element coordinates those NP, headed by [ferrum] and [aurum]. Regardless of the fact that [que] is a second-position clitic, for the purposes of linearization of syntactic structure, it must be the head of a &P. However, we have proposed, once coordinated, this complex NP is syntactically ‘flattened’, and taken as an opaque domain for agreement or extraction. For cases like this we proposed the term ‘que-coordination’, and argued that such a structure’s computational complexity *does not go beyond that of a regular string*[10], with the coordinated construction displaying a ‘flat’ syntax in the sense of Culicover and Jackendoff (2005). This means that the phrase structural template in (15) imposes too much structure on a coordination like (16b), a point that has also been made recently by Lasnik (2011) and which goes back as far as Chomsky (1963). We have proposed that this kind of coordination requires a simpler, ‘flatter’ structure, along the lines of (17) (see also Culicover and Jackendoff, 2005: Chapter 4 for a similar proposal, but generalizing *n*-ary branching):

17)

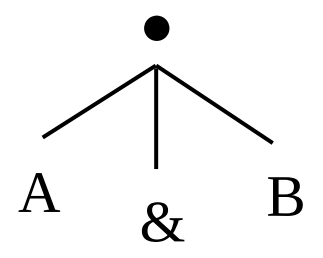

Differences in phrase structure are not exclusive to coordination, or to Latin. Consider the following situation involving modals and their interplay with negation[11]:

18) Du musst das nicht tun. (German)
You must$_{2Sg}$ that Neg do

That is *not* equivalent to English (19):

19) You must not do that

[10] In this context, ‘a X string’ (where X stands for ‘regular’, ‘context free’, ‘context sensitive’…) means ‘a string that belongs to a language that can be accepted by an X automaton’. A ‘regular string’ is thus a string that belongs to the set of strings that can be accepted by a FSA.

[11] Much of what follows is owed to Susan Schmerling, minus the mistakes this section might contain.

Rather, (18) is to be translated as (20):

20) You need not / needn’t / don’t have to do that

It has been argued (e.g., Itriadou and Zeijlstra, 2013) that the difference between German(-type) and English(-type) languages in this respect involves the relations of *scope* established between modals and negation:

21) a. Necessity(Negation(V)) → i.e., it is necessary that you don’t V (cf. 19)
b. Negation(Necessity(V)) → i.e., it is not necessary that you V (cf. 18-20)

But this is not it. Schmerling (1983a:12-13) has proposed that in English (at least), auxiliaries form a construction with the subject, from a Categorial Grammar stance that prefers ‘Item-and-Process’ grammars to Phrase Structure (derived from ‘Item-and-Arrangement’) grammars, following Montague (1970). Also following Montagovian semantics (see Montague, 1973), Schmerling investigated the possibility of auxiliaries belonging to categories (FC//IV/(FC/IV) and (IFC/IV)/(FC/IV)[12] which combine with nominative (FC/IV) subjects to form modified subjects, the latter belonging to categories FC//IV and IFC/IV. Schmerling argued that this possibility in fact accounted for phenomena that have been problematic for more traditional structures, including, for instance, discontinuous idioms and irregularities in auxiliaries’ paradigms. In formal terms, we could take Schmerling’s analysis as an early argument in favour of a mixed approach to syntactic structure of the kind we argue for in Krivochen (2015a), according to which the segmentation of a string *necessarily* involves neither (a) substrings of identical Kolmogorov complexity nor (b) identical computational dependencies between all members of the substrings. For instance, a segmentation of (18) should ‘conquer’ Aux-V before unifying that substring with Negation, so that Neg has the correct scope. If Schmerling’s analysis is on the right track, however, that analysis would not be valid for English, which would ‘divide’ so that Subj-Aux is a substring.

Moreover, Negation can configure a terminal together with Aux, as in (22):

22) John won’t arrive before 19:00

The epistemic auxiliary is grouped with the subject, but it does not go alone: Negation cannot be separated from the modal. Thus, a substring of (21) would be (23) (see also Schmerling, 1983a: 15 for a view from Categorial Grammar):

23) #John##won’t#

But bear in mind this analysis does not apply to items identified as ‘auxiliaries’ in all languages (e.g., Bravo et al. 2015, proposed that passive auxiliaries are in fact verbal morphemes, which would imply they are to be grouped with the verb rather than with the subject).

It is useful to take a look at (simplified versions of) the phrase markers a D&C approach, like MGG, proposes to account for scope variations:

[12] FC = finite clause, IFC = inverted finite clause, IV= intransitive VP (see also Montague, 1970)

24)

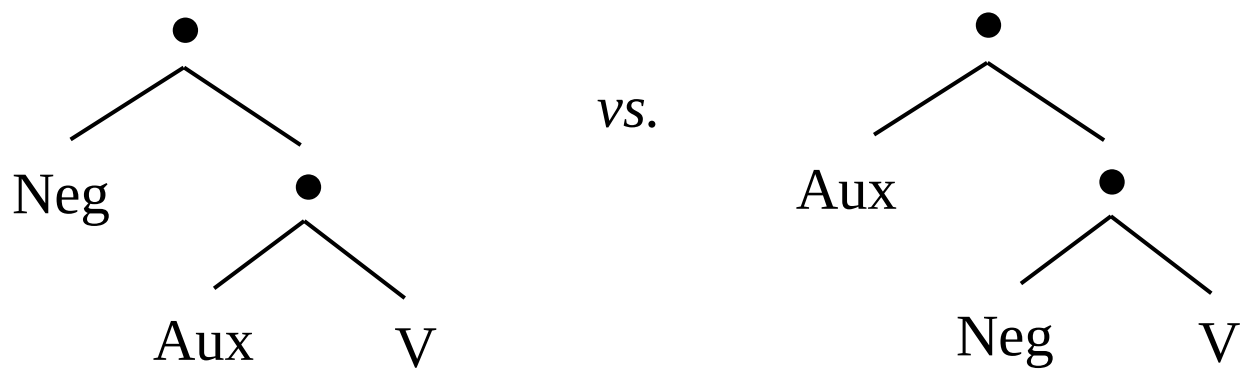

Semantic scope is in MGG a function of c-command relations, within a syntacticocentric approach to language. Itriadou and Zeijlstra (2013: 531) summarize this position as follows:

> (…) *the basic idea is that different modals are generated at different heights in the tree (some above negation, others below) and that differences between similar types of modals of similar quantificational force (like universal need to and must) are due to lexical idiosyncrasies* (…)

Related proposals include Neg adjoining Aux and covert LF-movement (see, e.g., Kosta, 2009 for a cross-linguistic perspective), in any case, the argument rests on semantic scope being read off a uniformly binary branching tree structure. In such a model, we would expect any head-to-head (i.e., terminal-to-terminal) movement to produce phonologically and semantically legitimate configurations…however, that is not the case. Whereas we do have 'need+not = needn't', and even weirder things like 'shall+not = shan't' (which is practically extinct in US English, however) –but, crucially, we have no *shalln't-, no 'will+not = *willn't', which would be expected under a strictly derivational approach and a free operation of Head movement (Internal Merge) in the syntactic component. There are also some creatures that resist the analysis on more fundamental ways, for instance:

25) John would rather walk

A uniform D&C parser (be it rule-based or constraint-based[13]) would be committed to a representation of (25) like (26):

26)

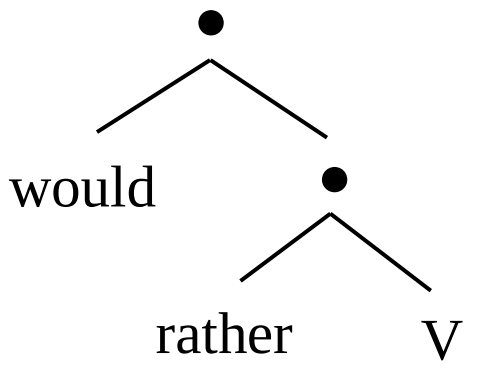

If (26) were right, then we would expect Neg, when represented in the tree, to adjoin the terminal that can materialize it, that is:

27) *John wouldn't rather walk

However, Neg applies to [would rather] as a whole (see Carl Baker, 1970: 169 for the earliest take on [would rather] as an idiom that we are aware of), that is:

28) John would rather not walk

[13] Hockett (1954) and Schmerling (1983b) present a distinction between Item-and-Process (pre-structuralist) and Item-and-Arrangement (Bloomfieldian and generative) theories. The considerations we have made with pertaining to D&C algorithms in theories of grammar affect both IP and IA theories equally, for they are both built over recursive processes.

At some point in the computation, when Neg enters the picture, then, [would rather] must configure a unit. The only way to capture this in a binary tree fashion complying with MGG requirements would be to have Neg either dominating the nonterminal in (26), or having the configuration in (29):

29)

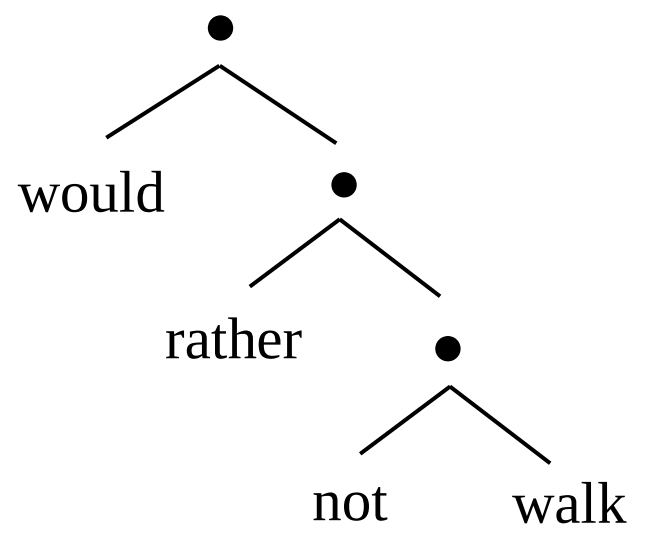

This option is problematic as well, for as Baker (1970, 1981) has pointed out, we can get things like

30) There is no one here who wouldn't rather walk[14]

If we assume that cliticization (at least in English and Spanish) is an instance of terminal-to-terminal (head-to-head) movement, and that such terminals have to be immediately adjacent (see, e.g., the *Head Movement Constraint* in Travis, 1984, according to which movement of X to Y cannot skip an intervening Z, for X, Y, Z, being terminal nodes), inducing structure from linear order would give us (31):

31)

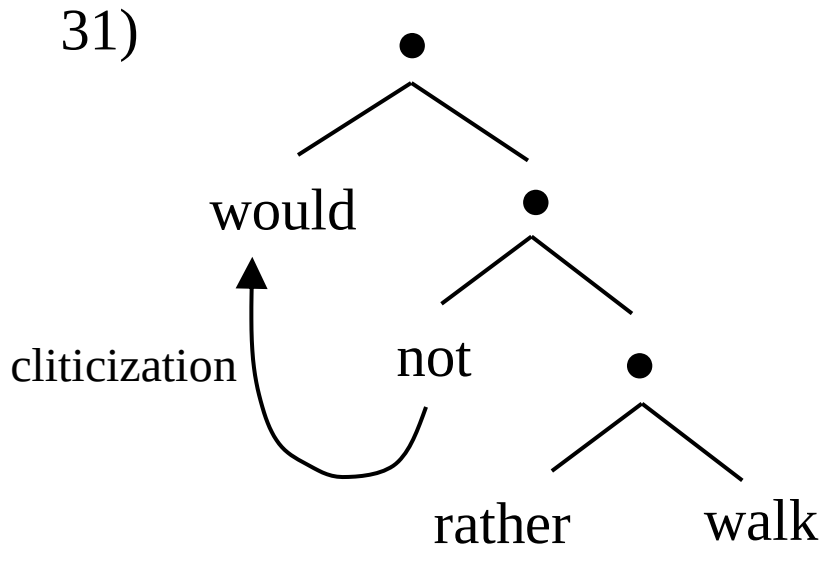

Yet, if we assume there is a Neg node between [would] and [rather], that only captures the linearization (by means of stipulation), but not the semantic scope: 'It is the case that no one would not rather walk' is *not* acceptable as a periphrasis of (30), according to the speakers we have consulted (southern-midlands UK varieties of English). Given the fact that c-command relations are read off directly at LF, it is not clear how a D&C proposal can handle the variability of Neg placement, if not by resorting to, for instance, some *transformation + deletion mechanism* that leaves Neg above [would rather] at the derivational point where interpretation takes place. A sorting problem of sorts (pun intended), but even if we phrase scope ambiguities as sorting problems (which we have seen are the textbook example of D&C), it is not clear what the criterion to follow for ordering would be (i.e., there is no obvious intrinsic cardinality among auxiliaries and negation) Such scope ambiguities do not hold with other Aux-Neg combinations, which only accept one scopal reading, and which are thus

---

[14] Susan Schmerling (p.c.) has pointed out to us that [would just as soon] behaves in a similar way:

i. I would just as soon not walk.
ii. *I wouldn't just as soon walk.
iii. There's no one here who wouldn't just as soon walk.

directly translatable to the uniform phrase structure template. In Spanish, for instance, word order is a major clue, and Neg appears next to the element it modifies, that is, next to the element it is to be 'divided and conquered' with. Thus,

32) No debes hacer eso
Neg must$_{2Sg}$ to-do that

can only mean 'You don't have to do that', with a standard (unmarked) intonational pattern.
In some varieties of English (particularly US Southern English, although it is not impossible for a UK Midlands speaker), an example like (33)

33) You may not sit here

is ambiguous between a deontic and an epistemic reading, which in turn depend on scope:

33') a. not(may(sit here) → deontic reading ('you are not permitted to sit here')
b. may(not(sit here) → epistemic reading ('it is possible that you do not sit here')

Of course the argument against a uniform D&C parser does not hinge on cases like (33) alone, whose ambiguity is arguable for some speakers (a minority, though), but (33) *does* pose a problem, in tandem with the other cases (and cases we have not considered, as well). The most compelling argument, without loss of specificity, is just that in English *the scope varies depending on the modal*. For example, [must] and epistemic [may] require Neg to have narrow scope, whereas other modals require it to have wide scope. We have to capture the fact that the scope is specified lexically, *somehow*: these are the kinds of situations in which a D&C algorithm turns out to be simply procrustean. If phrase markers are considered to be 2-D Hamiltonian graphs, uniformly binary-branching, these variable scope relations seem impossible to capture. There are –at least- two plausible ways to capture this behaviour:

a) Give up construal (i.e., give up the structure of VPs as composed by a Specifier, a V head, and an object Complement, and considered a syntactic unit), and assume elements are introduced freely in the phrase marker, as a function of semantic requirements

b) Give up the 2-D requirement for phrase markers *qua* graphs

In Krivochen (2015a: 554, ff.) we somewhat combined both proposals: we adopted a version of the 3-dimensional, Calder-mobile-like syntactic structure proposed primarily by Uriagereka (1998: 276–277), Lasnik et al. (2005: 35) and generalized the proposal to to *n*-dimensionality and *n*-ary branching. *n*-dimensionality, that is, the localization of structures in conceptual spaces defined by *n* coordinates, is compatible with semantic conditions insofar as conceptual structure can be mapped in *n*-dimensional vector spaces (Uriagereka, 2011; also Uriagereka's 2002 'warping' mechanism; more concretely, Zwarts and Gärdenfors, 2015 present an analysis of prepositions within a polar coordinates system): a phrase marker, if defined in terms of the location of its terminals in the conceptual space, can extend in more than 2 dimensions. However, it is possible that we also need to flatten that dimensionally rich structure dynamically in order to determine 'labels', basically, the 'name' nonterminals receive for the purpose of further computations (Lasnik and Uriagereka, 2011: 21; Saddy, 2016). Our proposal for the identification of a nonterminal involved a 'snapshot' of the Calder mobile in time, which yields a 2-D representation of an otherwise higher-dimensional structure. The generative engine that generates such higher-dimensional structures is, in our proposal, semantically driven, such that the phrase marker for a string (or sub-string) displays topological properties depending on the predication structures established within that string or sub-string. The proposal can be summarized as follows (Krivochen, 2015a: 560):

> *If a predicate is to have scope over a referential (either sortal or eventive) variable, the number of the predicate's coordinates in the mental working area properly contain the number of the arguments' coordinates.*

For instance, let us see what a predication relation over a sortal entity would look like in such a model:

(a) N = ($x$, $y$)
(b) A = ($x$, $y$, $z$)
(c) A(N) = A × B = ($x$, $x$') ($y$, $y$') (Ø, $z$)

And so on. For instance, if a category X (X ≠ A) was to take take (A(N)) as an argument, it would have to be defined in ($x$", $y$", $z$", $w$), introducing a further dimension represented by the $w$ axis. In this manner, phrase markers no longer yield *only* monotonic modification patterns. In this respect, Bravo et al. (2015) also present evidence from auxiliary chains in Spanish in favour of a non-uniform approach to phrase structure, such that a generative engine that generates a single kind of dependency (monotonic) is necessarily not sufficient to account for the syntactic and semantic behaviour of auxiliaries like progressive *estar*, passive *ser*, and perfective *haber* on the one hand; and aspectual *empezar* 'to start', *terminar* 'to finish', first position auxiliaries, root modals (*tener que*, *deber* 'to have to'), and verbs like *tardar* 'to take(time)' on the other. In that work, we argued that an auxiliary chain can mix CF and CS dependencies (i.e., embedding and crossing dependencies) *within* a structural description[15]. Let us briefly illustrate our point. Consider the abstract structure (34) and the corresponding example (35)[16]:

34)

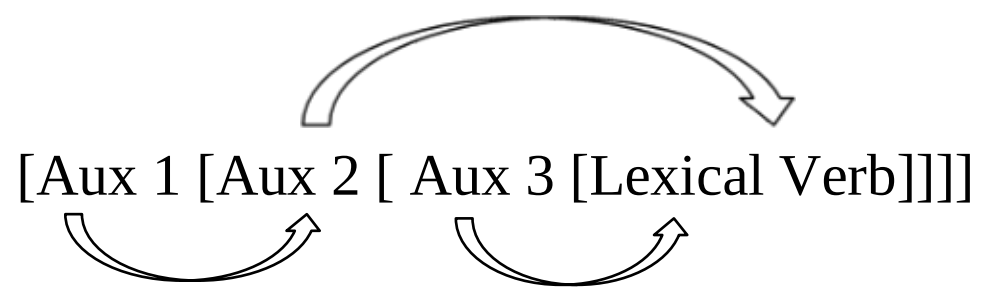

35) Ha tenido que ser ayudado por personal del centro.
*Has$_{3Sg}$ had to be helped by staff of-the center*
He/She has had to be helped by the center staff

Here we have an example of 'crossing dependencies' between Aux2 and the Lexical Verb, in which Aux 3 is transparent for the purposes of the modification relation Aux2(VP). However, within the auxiliary chain itself, Aux1 modifies Aux2 in a Type 3 (regular) dependency. A parser that assigned structural descriptions uniformly to all substrings of a language (a PSG-based parser, for example) is forced to assume modification relations within auxiliary chains are strictly monotonic, and discontinuous dependencies are not obviously captured by D&C algorithms when they are combined with modification under adjacency and which are sensitive to the nature of the elements involved and not just to their structural positions (as is the case for Aux3-VP, see fn. 12). The rest of the facts discussed in Bravo et al. (2015) also suggest that a dynamic model along the lines of Krivochen (2015a) is on the right track, at least in its essential *desiderata*.

---

15 This does not mean that, *descriptively*, strings belonging to a natural language L cannot be modelled by finite-state procedures (e.g., may → not → sit → here, in a very simplified form) for some very specific ends (e.g., unit segmentation). Uriagereka (2012: 53) claims that 'An exhaustively binary phrase-maker, none of whose branches symmetrically bifurcates, can be expressed in FS [finite-state] fashion'. We argue here (and in Krivochen, 2015; Krivochen & Schmerling, 2015) that such a modelling *cannot* capture the semantic properties of *every* string, and that an *explanatory* theory at the levels of syntax and semantics must go beyond the limits of a single level in the Chomsky Hierarchy, or else said theory will result procrustean.

16 There is a condition that seems inevitable for an instantiation of (34) to be an acceptable sentence: Aux3, which directly modifies V and is also structurally the closest one, must be a passive auxiliary. This is not a trivial requirement, as argued in Bravo et al. (2015), but we will not discuss it here.

All these considerations can be interpreted as *part* of an argument against a transparent syntax-semantics interface (i.e., assuming an *isomorphic* mapping) based on the recursive division of a string into pairs of substrings, which is the classic way in which MGG has approached phrase structure since the early days of the Standard Theory (see, e.g., Chomsky and Miller, 1963) to the Minimalist Program[17] (MP from now on).

Contrastingly, Schmerling (1983a) proposes a dynamical approach to categorization, based on a modified Montagovian approach:

> *The independent stipulation of the scope of the negation for different modals indicates that the negative modals are basic expressions.* Schmerling (1983a: 16)

She goes on to argue that irregularities are better tackled if what counts as a 'basic expression' is not fixed beforehand following aprioristic syntactic templates, which is in consonance with the mixed dependencies model we have been arguing with. If an argument along these lines is correct, then not all substrings have the same length, and it is thus not obvious how the *divide* algorithm would work, if it is kept insensitive to semantic considerations (idioms also constitute a powerful argument against uniform phrase structural descriptions, see Schmerling, 1983a for discussion and analysis in terms that are germane to those used here). But if the chunking procedure is made semantics-sensitive (i.e., if *cycles* are not fixed beforehand), then the variability of chunk size and complexity makes us question the apparent computing time advantages D&C algorithms offer: the process cannot always be carried out with configurational information only, for we have seen examples of ambiguous configurations like [may not], and D&C algorithms implemented by deterministic automata (of the kind assumed for sorting, which can be extended to transformational grammars) are not obviously sensitive to the possible interpretations of those configurations. Turing (1936: 232) claimed that

> *When* […] *a machine reaches one of these ambiguous configurations, it cannot go on until some arbitrary choice has been made by an external operator.*

The thing is, we cannot predict *where* or *when* ambiguous configurations will appear, which is to be expected if semantic interpretation is *not* a linear function of phrase structure. Considering this, if we can identify patterns of D&C-computable substrings, including *iteration*, *que-coordination*, and possibly some instances of *adjunction*, then the parser must be not only semantics-sensitive, but also

---

[17] In this respect, Lexical Functional Grammar (LFG) appears to have an advantage, since

> *All syntactic phrase structure nodes are optional and are not used unless required by independent principles* (…) (Bresnan, 2011: 115)

However, an explicit formulation of those 'independent principles' that govern phrase structure has not yet been devised (the same can be said of Culicover and Jackendoff's 'Simpler Syntax').

able to identify the D&C computable substrings, solve them, and leave the rest for other kinds of algorithms. Or, as we have argued, approaches to natural language processing are not algorithms in the strict sense of the word at all (despite the relative frivolity with which the word 'algorithm' is sometimes used[18]). In other words, *the linguistic parser cannot be algorithmic*, in the strict sense of the word (see Wegner, 1997; Milner, 2006; Goldin and Wegner, 2007, for developments of the idea of *interaction-based computation* as the basis of cognitive operations as opposed to the narrower *function-based computation*).

It is essential to distinguish the original CTT from its extensions, which are very common in the literature. Thus, for instance, Fitz (2006) proposes and discusses the so-called Physical Church-Turing thesis, according to which a function is computable by a physical system *if and only if* it is TM-computable (see also Deutsch, 1985 for a view that is related, although more influenced by quantum mechanics). This claim amounts to saying that all computations within physical systems are function-based, which in fact requires independent formal proof. Deutsch (1985: 3) explicitly formulates the physical CTT principle as follows:

> *I can now state the physical version of the Church- Turing principle: 'Every finitely realizable physical system can be perfectly simulated by a universal model computing machine operating by finite means'.*

However, Deutsch himself acknowledges a version of the CTT that only deals with function-based computation while not providing a formal proof that every finitely realizable physical system is actually function-based (strangely enough, no citation information is provided for this formulation, which is attributed to Turing):

> […] *according to Turing,*
> *Every 'function which would naturally be regarded as computable' can be computed by the universal Turing machine.* (Deutsch, 1985: 3)

Deutsch advances on the implementational concerns pertaining to the CTT, by pointing out that:

> *I propose to reinterpret Turing's 'functions which would naturally be regarded as computable' as the functions which may in principle be computed by a real physical system. For it would surely be hard to regard a function 'naturally' as computable if it could not be computed in Nature, and conversely* (Op. Cit., 3)

The 'conversely' part is the one we are not so sure about: along with Wegner (1997), Milner (2006), among others, we claim that there are computations that occur in 'Nature' (e.g., the development of L-systems; interactive-based computation in general, including, for instance, the process of deriving implicatures in linguistic utterances) which are not sequential or function-based, for we either have non-sequential rewriting or external factors influencing the derivation at points which cannot be fixed *a priori* (see, e.g., Wilson and Sperber, 2004: 615).

Copeland (2002) discusses Deutsch's (1985) physical extension of the CTT, claiming that

---

[18] For instance, "minimal search gives you ***a kind of obvious*** *algorithm* for which piece of the structure is relevant to further combination – labelling" (Noam Chomsky, in discussion with Cedric Boeckx, 2009. Our highlighting). No explicit formulation of that 'kind of obvious' algorithm is provided. Consider that, following Rogers (1987), for instance, an algorithm is an effective method that can be expressed within a finite amount of space and time and in a *well-defined formal language* for calculating a *function*, the above quote cannot be taken in all seriousness.

*The notion of an effective method played an important role in early debates about the foundations of mathematics, and it was sufficiently clear to allow Turing, Church, and others to recognize that different formal accounts gave alternative modellings of the notion. Their notion was certainly not that of a 'finitely realizable physical system'*

but himself proposes a version of the CTT that is stronger than Deutsch's: all *effective computation* is carried out by a TM. Thus, all effective computation is function-based in this view, and can be modeled by means of a TM:

*There are various equivalent formulations of the Church-Turing thesis. A common one is that every effective computation can be carried out by a Turing machine.*

In Copeland's article, 'effective' is not (at least explicitly) equivalent to 'function-based', therefore, we can safely say that, despite the possibility that this is an independent thesis from CTT, it is certainly not equivalent to the original CTT (which makes no reference to *effective computation*). It is crucial to point out that Turing's object in his seminal (1936) paper were computable numbers, but he claims his theory can be extended '*to define and investigate computable functions of an integral variable*' (1936: 230). Crucially, a TM (and thus, any other model of computation that claims to be formally equivalent to a TM) is at every step determined by its *configuration*, which includes the symbol that is currently being read $\mathfrak{S}(r)$ and a condition $q_n$, from the set $\{q_1, \ldots q_n\}$ of the possible states of the machine (1936: 231), what Turing calls an 'automatic machine' (or α-machine). However, Turing himself concedes that

*For some purposes we might use machines (choice machines or c-machines) whose motion is only partially determined by the configuration* […]. *When such a machine reaches one of these ambiguous configurations, it cannot go on until some arbitrary choice has been made by an external operator. This would be the case if we were using machines to deal with axiomatic systems. In this paper I deal only with automatic machines* (…)

For our purposes, it is essential to keep in the foreground the fact that the formalization in Turing (1936) is aimed at automatic machines, since this helps us grasp the true scope of CTT and the impact it has had in formal linguistics, particularly linguistic theories of generative orientation.
To summarize our discussion in this section, given a string *uvwxy*, there is no way of knowing beforehand whether an arbitrary substring is D&C solvable, contrary to what would be predicted by a uniform approach to syntactic structure. This is so, we argue, because *natural language strings are not computationally uniform*.

## 4. On the transformational component:

Transformational models are by definition *polystratal* frameworks, in which there are 'hidden levels' (Culicover and Jackendoff, 2005: 16) where elements are added, moved, or deleted by means of transformational rules[19]. These levels proliferated in the Standard Theory (where we had a Base Component including the Lexicon and a set of Phrase Structure rules, Deep Structure, and Surface Structure, the last two related by means of a Transformational Component) and the Government and

[19] LFG, Representation Theory, and Simpler Syntax all feature more than a single level of representation, but since these levels are built using different elements, it is not correct to say that one has been mapped onto another via transformations: they are all equally 'base generated' and related via mapping rules and/or constraints.

Binding theory (featuring D-Structure, S-Structure, Logical Form, and Phonetic Form, other parts of the theory configuring *modules* rather than levels of representation). Such levels required mapping functions that were reversible, insofar as we could get from a level L' to the one it derived from L if we knew the specific transformational rules that mapped L onto L'. Even in the MP, which argues for derivational and representational economy, there are still transformations (although under a different guise; mainly Move-α / unbounded Internal Merge, be it overt or covert –at the level of LF) applying to features or bundles of features; therefore reconstruction effects are ubiquitous (applying to overt A/A' chains as well as quantifier raising; see Lebeaux, 2009: 7, ff.). Thus, there is a pre-transformational structure that must be re-built: this implies a sorting mechanism, which is aided by an aprioristic theory of possible landing sites (in both the Barriers and the Phases frameworks; see Chomsky, 1986b, 2000 respectively). The copy theory of Movement and the operation Spell-Out has in fact revamped the process of Equi-deletion, which implied NP deletion under identity. Berwick (1984), explaining earlier results by Peters and Ritchie (1973), correctly points out that the reconstruction of a pre-transformational structure or level of representation (i.e., Deep Structure, in the Standard Theory and its extensions) greatly increases parsing time, up to EXP (exponential time, a function of the form $n^x$). However, he claims that in GB, 'deep structures (…) need not be built at all to test grammaticality' (Berwick, 1984: 189). That is inaccurate for D-structure and deep-structure understood as pre-transformational levels (but is trivially true insofar as D-structure was claimed to be different from deep-structure, see Chomsky, 1986a, b): at least two modules of the grammar required direct reference a pre-transformational phrase marker in order to test grammaticality, namely, *Theta Theory* and *Binding Theory*. Let us focus on the latter, for it provides us with compelling arguments that are valid even nowadays, when GB is considered dead and buried.
Binding Theory (BT) was conceived of as a 'module' (a set of well-formedness principles) of the theory of grammar which determined the distribution of three kinds of elements: pronouns, anaphors (reflexives and reciprocals), and R-expressions (proper names and determiner phrases more generally). This 'module' consisted of three principles (Chomsky, 1981, 1995; Kosta, 1992):

36) Principle A: an anaphor is always bound in its governing category[20]
Principle B: a pronoun is always free in its governing category
Principle C: a referential expression (R-expression) is always free

BT presents two kinds of problems: on the one hand, the very formulation of the principles (see Lasnik, 1997 for some discussion, we will not get into these problems here); on the other, the derivational timing for the application of these well-formedness conditions.
The latter problems were tackled, among others, by Lebeaux (2009). Interestingly, and even well into Minimalism, the necessity for the reconstruction of pre-movement structures for the purposes of BT and semantic interpretation is still a problem. Consider for instance a structure like (37):

37) John doesn't admire Mary, and he doesn't admire Bill. But now [himself, John admires]

Let us focus on the clause with the reflexive (the others setting the context). If we divide and conquer, we are left with the following tree representation:

[20] Oversimplifying a bit, let us say that the governing category for α is 'the minimal I(nflection)P / N(oun)P containing α, the lexical head that selects it, and its antecedent'. Governing categories captured the notion of *locality* in GB before *barriers* were proposed.

38)

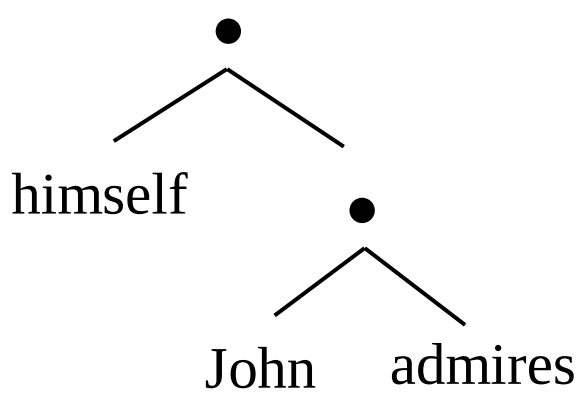

However, such a representation clearly violates Principle A, since the anaphor [himself] c-commands its antecedent [John]; thus, (37) should be ungrammatical, which it clearly is not. The crucial thing here is that there is *no way* to salvage this structure *unless* we assume that there is a transformation that has applied, displacing the anaphor from a pre-transformational (or base-generated) position in which it was c-commanded by [John], to its surface position: Principle A then applies to this pre-transformational structure, and the well-formedness of the phrase marker 'carries over' throughout the derivation. That is, (38) has to be replaced by (39) as the corresponding structure for the bracketed clause in (37):

39)

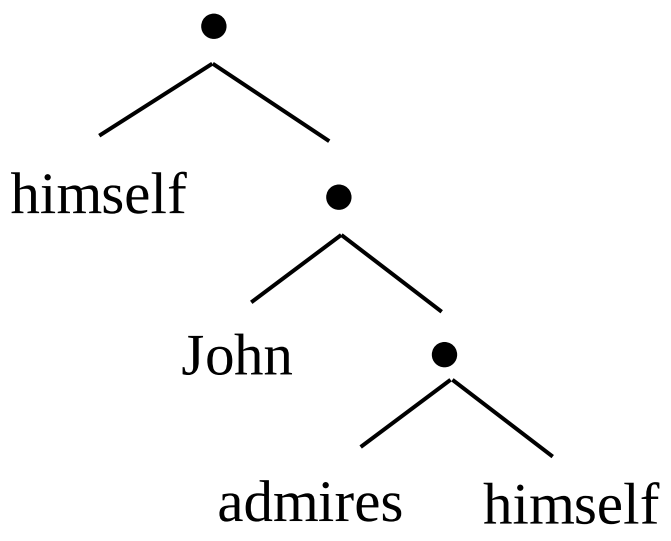

Even in this simplified structure, we need to add a nonterminal in order to create a structural position where [himself] would comply with Principle A. But we cannot get to that structure via reverse-engineering the LCA, and there is in fact no upper bound on the number of nonterminal nodes we could have proposed between [John] and [himself]: once we accept the existence of transformations, we cannot limit their weak generative power unless we increase the size of the grammar by means of the addition of constraints. Ross (1970: 149) insightfully claimed that '*it seems to be the case that even in apparently simple sentences, the transformational mapping between deep and surface structure is extremely complex -far more so, in fact, than has previously been thought*'. This is true of the base phrase markers corresponding to gapping, sluicing, etc., but also of the operations that reconstruct pre-transformational phrase markers in order to avoid violations of BT principles, either by recovering a D-structure or by re-building it at the level of Logical Form. Regardless of where it happens, at some point we need to build a phrase marker in which structural relations are either those in (39) or reducible to (39) (e.g., if we assume intermediate 'functional' phonologically empty nonterminals that then are ignored for linearization purposes). Otherwise, the theory incorrectly predicts that (38) is ungrammatical. Reconstruction (that is, the process of tracing the transformational history of a syntactic object) is widely assumed as a test for both A(rgumental)- and A'- (non-argumental) movement and binding properties of quantifiers (Lebeaux, 2009: Chapter 2; Ruys, 2011; Fox, 1999, among many others). Fox (1999: 158) claims that scope reconstruction of quantifiers is determined by the syntax, '*the structures that serve as the input to semantic interpretation (the structures of LF) determine whether or not there is scope reconstruction.*': reference to a pre-movement (i.e., pre-transformational) structure is inevitable. But such a set of procedures (i.e., reconstruction) needs to adhere to some guidelines, for instance, the kind of structures that are legitimately base-generated (and are thus plausible candidates for being 'reconstructed'), and an explicit description of the permissible transformations over such structures. Reconstruction must therefore proceed in a D&C way.

But, is there anything in a post-transformational structure that allows us to limit the (in principle) unbounded power of reconstruction (i.e., a gross overgeneration of possible pre-transformational structures)? Yes, there is; constituents leave derivational crumbles, in a Hansel and Gretel kind of syntax: *traces* / *copies,* deleted at PF but visible at LF (Nunes, 2004: 11; Corver and Nunes, 2007: 2). The mechanism works as follows (in terms that are widely accepted within MGG):

40) a. Copy a syntactic object SO, yielding SO'
    b. Merge SO' so that it extends the phrase marker
    c. Form Chain {SO', SO}

In this framework, a Chain is a pair of occurrences of a constituent in a phrase marker. Thus, both occurrences of [himself] in (39) constitute a Chain.
The notion of Spell-Out applied to chains yields selective materialization, for not all occurrences of a SO receive phonological exponents. Thus, there are copies that are there in a phrase marker for the sole purpose of complying with some internal syntactic requirement (e.g., related to the impenetrability of some domain) and do not receive a phonological exponent. These intermediate copies, which are in principle no different from traces in strictly formal terms (as we showed in Krivochen, 2015b), and the indexing that takes place in order to identify occurrences of a SO as members of a chain, has to happen in both Copy+deletion frameworks (GB) and in Lexical insertion+indexing frameworks (ST). The idea is that SO and SO' have to be linked somehow: even if there are no graphical indexes, identity has to be encoded in some manner.
Berwick (2015) compares Minimalist grammars with HPSG/LFG types of grammars, in which displacement is encoded via *slash-features*[21]: VP/NP is to be read as 'A VP that contains an NP gap'. This is relevant for the present discussion because *slash*-based grammars encode gaps at each derivational point, and thus the size of the grammar expands. Berwick analyses the representation such a grammar would assign to a multiple-gap construction like (41):

41) [Which violins]$_i$ are [these sonatas]$_j$ difficult to play $t_j$ on $t_i$?

The *slash*-based representation of (41) is, according to Berwick (2015: 7), the following:

42) CP → *wh*-$NP_1$ (*which violins*) S/$NP_1$ ; S/$NP_1$ → $NP_2$ (*these sonatas*) S/$NP_1NP_2$;
    S/$NP_1NP_2$ → (*pro*) VP/$NP_1NP_2$; VP/$NP_1NP_2$ → V $NP_2$/$NP_2$ PP/$NP_1$; $NP_2$/$NP_2$ → ε;
    PP/$NP_1$ → $NP_1$/$NP_1$; $NP_1$/$NP_1$ → ε;

He proceeds to argue that there are intermediate constituents that play no role in the filler-gap relation. Notice, however, that for an example like (41), (42) presents us with a way to reconstruct each derivational step: thus even though the size of the grammar increases exponentially, each context-free rule exhausts the description of the relevant syntactic object at every derivational step.

[21] It must be noted that, while the notation is the same, the meaning of A/B is quite different in HPSG and in Categorial Grammars, where it originated. Müller (2013) considers that *modern* versions of categorial grammars (which present significant differences with Ajdukiewicz's original formulation) are equivalent to Minimalist Grammars (in the sense of Stabler, 2001) and HPSG. Interestingly, the fraction notation of original categorial grammars can make headedness explicit in the same way the '<' '>' notation of Stabler for phrase structure trees.

Let us see how a GB grammar would perform here. Using Chomsky's (1986b) framework[22], the relevant phrase structure representation would be (43):

43) [Which violins]$_i$ are$_k$ [these sonatas]$_j$ $t_k$ difficult [PRO$_l$ to play $t_j$ [Op$_i$ on $t_i$]]

Notice that two elements proliferate here: indexes and empty categories. Being committed to strictly binary branching and assuming the *Barriers* framework, MGG has no choice but to insert an empty operator in the PP adjunct [on _], otherwise, the NP$_i$ trace would not be *properly governed*, the representation would violate the Empty Category Principle (a UG principle ruling the distribution of traces and their antecedents) and thus it would be ill-formed. The null subject PRO for the infinitive [to play] is also a theory-internal requirement, for without it, the Extended Projection Principle, which requires all clauses to have a subject, would be violated. We also have T-to-C movement (which moves [are] before the subject), and ubiquitous indexes relating gaps and fillers. The indexes *i* are even *nested*, that is, we get [NP$_i$…[Op$_i$…$t_i$]] (see Berwick, 1984: 191 for some discussion of index nesting): we need to get into the adjunct (which is otherwise impenetrable; see Chomsky, 1986b; Huang, 1982, *inter alios*) in order to get the right indexing. As a description, (43) arguably works, but the process of actually reconstructing the filler-gap relations is not so straightforward. How big, then, is the grammar? Berwick (1984) argues that with the replacement of Equi-deletion by PRO, the space is shrunk from exponential space to polynomial space. However, it is not at all clear that the advantages of lexical insertion and indexing over copy and deletion are real: after all, just like there are unbounded deletion possibilities, nothing prevents us from assigning indexes freely, other than stipulations over the Numeration (e.g., determining beforehand that syntactic objects $SO_1$, $SO_2$, …, $SO_n$ selected from the Lexicon to be part of the Numeration used to derive a particular sentence, will share referential indexes).

The situation is not much better with *copies* instead of traces. PRO is still maintained in the theory (except for the Control-as-Movement proposal of Hornstein, 2001, but in that case, we multiply indexes as well as features, since we have to encode theta-theory into feature matrices), but now we have to consider chains of occurrences as relations between copies of a syntactic object of arbitrary complexity. Summarizing discussion in Krivochen (2015b), we can say that Chomsky (2000) claims that the inclusion of traces in a derivation violates the Inclusiveness Condition (which bans the inclusion of elements that are not present in the Numeration during the derivation), and therefore, they are to be replaced by Copies. However, the operation Copy also introduces new elements in a derivation, provided that the copies are not present in the initial Lexical array or Numeration; therefore, copies also violate the Inclusiveness Condition if this condition is to be understood strictly: information cannot be deleted or lost, but it should also be impossible to add information (in the form of syntactic terminals, for instance) that is not present in the Numeration, including distributional specifications for copies[23]. If an element of arbitrary complexity is copied and then Merged, there is no reason to believe that element was present in the Numeration (unless we assume a massive amount of looking ahead that allows the Lexical Array/Numeration to see what is going to be the output of the derivation and thus have all needed elements, including copies, ready beforehand), it is therefore treated as a whole new element: after all, until we get to the interface levels, we have no possibility of establishing a referential connection with an object already introduced in the derivational space, except under special stipulations which depart from the simplest scenario and require justification. Conversely, if copies were indeed present in the NUM, the operation Copy would be superfluous, since there would be nothing to copy: all usable elements would be already predicted (somehow) in the NUM. Müller (2013: 940) even suggests that, when comparing MGG and HPSG accounts of apparent 'remnant movement' in German, HPSG's 'argument composition' variant '*would appear to need less theoretical apparatus*' (but he says nothing about grammar growth, which is Berwick's main objection). An alternative to both phrase structure-based approaches is highly desirable, for their respective shortcomings follow from basic assumptions regarding the computability of syntactic structure under strong uniformity assumptions (even though the specifics vary between frameworks, the basic D&C template seems to be 'universal').

[22] Truth be told, there are no big differences in the representation of multiple-gap constructions between GB and MP. Derivationally, it could be argued that Sidewards Movement (Nunes, 2004) has some advantages over unrestricted Move-α, but the empirical coverage of the theories is practically the same. Moreover, the Sidewards Movement operation was devised for Parasitic Gaps (e.g., 'That is the book I filed *t* without reading *t*'), multiple-gap constructions being addressed only as an extension of the original formulation.

[23] Stroik & Putnam (2013: 20) express a similar concern: "*To "copy X" is not merely a single act of making a facsimile. It is actually a complex three-part act: it involves (i) making a facsimile of X, (ii) leaving X in its original domain D1, and (iii) placing the facsimile in a new domain D2. So, to make a copy of a painting, one must reproduce the painting somewhere (on a canvas, on film, etc.), and to make a copy of a computer file, one must reproduce the file somewhere in the computer (at least in temporary memory).*" The identity of such temporary memory is not addressed in MGG.

It is crucial to notice that there is nothing in either the *trace* or the *copy* theory of movement that prevents unbounded deletion, and, moreover, *phase-based* versions of the *copy theory of movement* actually make things more problematic, for the number of copies is multiplied. Consider that (at least) transitive *v*Ps and CPs are *phase heads* in current generative models (Chomsky, 2008; Gallego, 2010; Citko, 2014), meaning they define domains which are impenetrable for purposes of operations triggered by either the immediately higher phase head (in Chomsky's 2001 version), or any head outside the phase maximal projection (in the early 2000 version of the theory). Let us diagram the situation:

44) For X and Y phase heads, Z a non-phase head

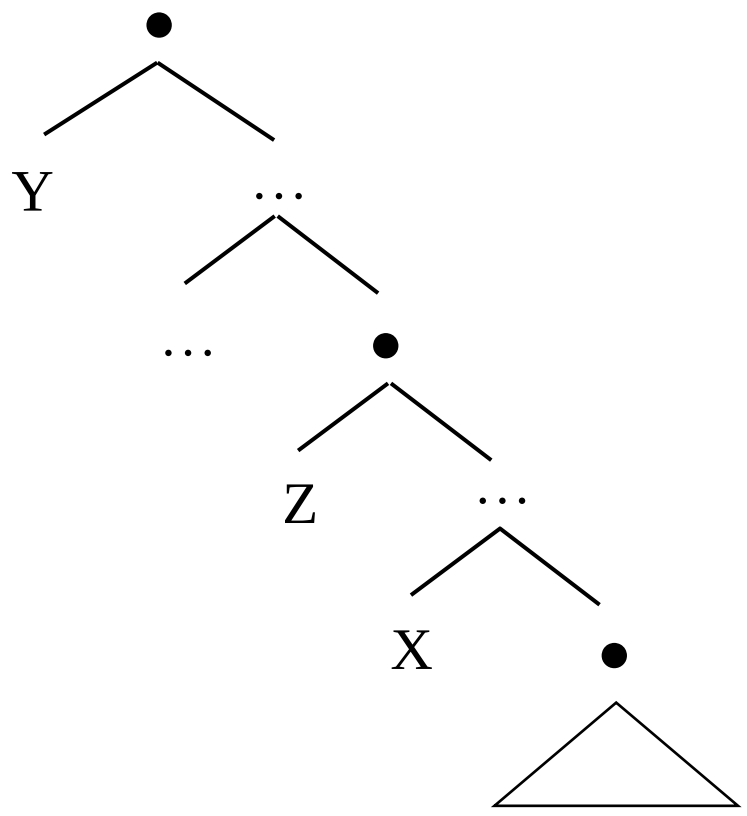

In either version of the theory (2000, 2001), the sister node of X is impenetrable by further computations (including transformations targeting a member of X's sister nonterminal, basically Agree and Move in MGG). This means that, for a syntactic object to be extracted from the complement of X or anything dominated by that nonterminal, that element must move first to a position *outside* the c-command domain of X, what is technically referred to as the *periphery* of X. Any phase projection (i.e., any projection of a phase head) thus contains a copy of the target of displacement in the periphery of the phase head, in an *outer specifier position* which plays no role in interpretation, being motivated only by an intra-theoretical requirement. Let us see an example:

45)

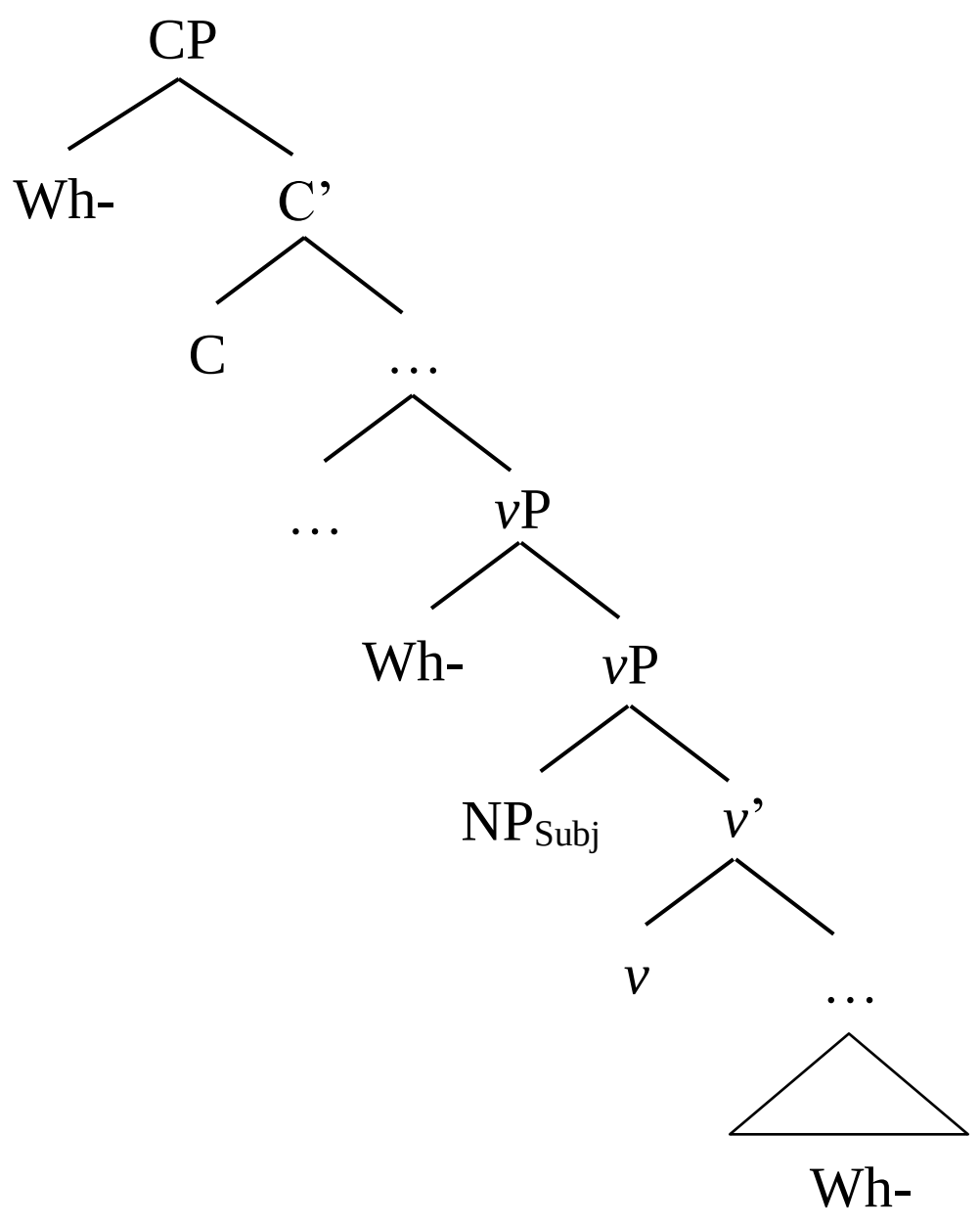

In a language like English, of the SVO kind, the lowest copy would get thematically interpreted, whereas the highest copy would yield interpretative effects including presuppositions. However, the copy in the phase head *outer specifier* has no interpretative impact. Moreover, since the system does not know when the next phase head will be introduced, or more fundamentally, how many derivational steps we must wait for the final landing site for the Wh-word to appear. The system is thus required to maintain a copy of the displaced constituent active in a workbench that can be readily accessed. If one assumes a DTC approach to computational complexity, as is the norm in MGG (see, e.g., Marantz, 2005; also Berwick, 2015), then the *m*-configuration of the automaton in charge of processing the structure (using Turing’s 1936 terminology) must include, in case it is necessary, the specification of all elements to be carried over throughout the derivation, from zero (no displacement) to indefinitely many, only limited by the number of movable elements in a string. Typological factors would determine the amount of possible landing sites (e.g., consider the availability of scrambling + Wh-movement in German; or multiple Wh-fronting in Japanese; see the articles in Boeckx and Grohmann, 2003 for discussion and examples). Using the same notation as in (42), for *n* moved constituents within a *v*P phase, the specification of that *v*P must include:

a) The outer specifier positions occupied by displaced elements (which do not get interpreted)
b) The copies / traces left behind (which would get thematically interpreted)

Thus, using the notation in (42), the *v*P in (45) must be specified as (46):

46) *v*P → Wh-, *v*P/Wh-

In traditional CG notation, *v*P/Wh- means that the nonterminal thus defined is to be combined with a Wh- to yield a complete *v*P (Lyons, 1968: 227, ff.). However, in more recent HPSG terms, (46) means that the *v*P contains a gap of the Wh- kind (see, e.g., Gazdar, 1981: 159). Interestingly, from a phase-theoretical stance, *we need both*. Why? If we have no copy of the Wh- element in an outer specifier position (i.e., if the *v*P does not combine with an instance of the relevant Wh- element), then it cannot be displaced out of the relevant phase, for phases are impenetrable domains. Moreover, we have to encode the fact that the element in the periphery of *v*P corresponds to an element that is referentially identical to a gap within that *v*P. The specifications we need at each derivational point are thus more complex than they are in MGG’s non-transformational counterparts. We are unaware of works that deal with phase-based Minimalist grammars from a computational point of view (note that Michaelis, 2001; Stabler, 2003, 2013; Berwick, 2015 and others do *not* implement the phase framework), there being a gap (pun intended) between linguistic theory and mathematical modelling.

Aided by the axiomatic character of binary branching in the MP, D&C algorithms (say, of the *merge-sort* kind) could tackle the reconstruction of the phrase marker given an *m*-specification like that in (46) plus additional stipulations over the properties of a tree and relations among elements of that tree (e.g., c-command(*x*, *y*)). The process of interpreting copies *requires* a sorting mechanism that, given a domain (say, a *phase*), can link copies in the periphery of a phase head (be it the moved object's final destination or not) with lower copies within that domain. The process is not as simple as it seems, because, unlike HPSG entries, copies in Minimalism move their way up a syntactic tree by checking / valuating and discharging features that form lexical items and cannot be interpreted by the semantic or the morphophonological systems (so-called 'uninterpretable' or 'formal' features; see Chomsky, 1995: 276, ff.). This means that in any configuration, no two copies of a syntactic object are defined by the same feature specification if operations are indeed driven by the necessity to valuate/check features (*contra* Rizzi, 2004). Identity is not possible in a system that admits feature checking relations, since the feature matrix of an element varies as the derivation unfolds and copies are (Internally) merged in places in which they can valuate and erase uninterpretable features. The interface systems cannot establish a dependency between two objects, say, α and β, as in (47):

47) α = {*i*-$F_1$, *i*-$F_2$, *u*-$F_3$, *u*-$F_4$}
β = {*i*-$F_1$, *i*-$F_2$, ~~*u*-$F_3$~~, ~~*u*-$F_4$~~}

Where *i* = interpretable, *u* = uninterpretable.

because there is nothing inherent to α and/or β that suggests they are linked. Any link should be encoded as a diacritic of sorts (e.g., a referential index that is carried throughout the derivation). Let us assume the following scenario:

48) $F_1$: Categorial feature (e.g., V, N)
$F_2$: ϕ (Agreement features)
$F_3$: Case
$F_4$: Wh-

In the first-Merge position, the element has the feature matrix α, whereas after checking Case with the *v* or T heads and movement to Spec-CP, the matrix is β. We see that the identity criterion is not met: the feature matrices are different.

Is there an alternative to PSR-based accounts of non-local dependencies and displacement? Taking into consideration that, along the lines of cognitively based grammars, phrase markers and operations over such markers have a mental reality, we have proposed one such alternative, based on transformations over topological spaces in which derivations take place. This does not mean it is the only one (see Müller, 2013; Sag, 2010; Martin and Uriagereka, 2014 for equally consistent alternatives, which diverge to different extents from what we have presented so far)[24]. What must be

[24] There have been recent attempts to show (successfully, in our opinion) that monotonic, uniform phrase marker-based grammars are in fact notational variants of each other (Müller, 2013) or that, for example, if the functional structure of the MP was fully fledged out, the MP would be equivalent to a Categorial Grammar (Sag, 2010). Those papers are worth looking into, as is the extent to which a non-uniform proposal, which includes some insights from the mathematical analysis of non-normal grammars (see, e.g., Patel et al., 2015), is also 'reducible' to a notational variant of one of the previously mentioned frameworks.

borne in mind is that it is a consistent alternative that is focused on the dynamical properties of the topological space in which derivations are hypothesized to take place, and that it corresponds to a cognitively real space (as opposed to a purely formal space where derivations as proofs are developed independently of time, as with MGG, HPSG, and LFG). Our proposal in Krivochen (2015b) extended and developed previous work in Krivochen & Kosta (2013), applying a type-token approach to displacement mechanisms. To summarize this proposal, once we have a free, dynamic structure building operation that relates *n* elements at each artificially quantized step, 'long distance' relations depend on a notion of 'distance' that we define, and how to overcome it. Our proposal is that the Lexicon contains *types*, and those *types* are instantiated as *tokens* in different structural positions (see also Martin and Uriagereka, 2014: 175 for a somewhat different perspective). In our proposal, a sentence like

49) Who thinks who believes Bob?

(which we analyzed in Krivochen, 2015b), contains two *types* of [who], each of which corresponds to a different entity. Each of those *types* is instantiated *only* in semantically relevant positions, such that *every position in which a token appears must contribute to the interpretation of that token once the chain is collapsed* (Uriagereka, 2011; Martin and Uriagereka, 2014). That is, in consonance with LFG desiderata, a WYSIWYG kind of structure: there are no positions licensed by strictly intra-theoretical notions like 'phase head' or 'chain uniformity' (for arguments *against* the notion of chain uniformity, for instance, see Kosta & Krivochen, 2014). Of course, we need to relate the tokens (the 'collapse' of a number of occurrences into 'unique positions' proposed by Martin and Uriagereka, 2014: 180), and we have provided (2015b: 30) a principle to this end:

*Token-Collapse*

*Let S be a set {α, β, …, n} of arbitrarily complex tokens in positions P within a derivational workspace W. An Interface Level IL establishes a dependency between the members of S iff:*

*a. The members of S are mutually disconnected*
*b. No two members of S belong to the same domain D, and there is no syntactic object SO, such that SO ∈ D and SO is logically equivalent to a member of S for interface purposes*
*c. The members of S are structurally identical as far as format is concerned* [i.e., they are both terminals or both nonterminals]

Let us see another example:

50) Who did you believe Bill to have seen?

If the thesis is to hold that ECM constructions involve CP deletion (i.e., they are bare TPs), the timing of movement in relation to deletion timing is problematic: orthodox accounts need the intermediate landing site for [who] at Spec-CP before its deletion for Subjacency reasons, more recently subsumed to the Phase Impenetrability Condition (Müller, 2011: 13). However, a type-token account derives the sentence using a single [who] type (details of [who] displacement have been provided above), with two token instantiations: Compl-V and Spec-C (to use familiar X-bar terms). Since [Bill] is not a token of [who] (they do not share a *denotatum* or have the same referential properties), it is not an intervening element for token-collapse purposes.

The above Token-Collapse principle assumes a graph-theoretic approach to Chosmky-Greibach normal phrase markers, which is convenient formally, but not obviously so neurocognitively (or, using Marr's 1982 terms, 'implementationally'). More recently, we have formulated an alternative of this approach within a field-theoretic topological approach to syntactic structures as cognitive entities. Under the assumption that derivations are topological operations over a metrizable space, a relation between a number of elements is defined as the interference pattern that those elements, which are themselves perturbations of a field (see also Piatelli-Palmarini and Vitiello, 2015 for a related perspective also based on field theory): all of these elements can be readily formalized using well-known equations in Quantum Field Theory. This perspective follows from a field-theoretic approach to the Lexicon and the assumption that cognitive operations are topological in nature, as opposed to strictly formal operations captured by algorithmic procedures of the D&C kind. The warping mechanism proposed in Uriagereka (2002) requires phrase markers to have topological reality, and we have expanded on that proposal within a more strongly topologically oriented theory: phrase markers as topological spaces 'warp' or fold onto themselves, self-intersecting at a point or set of points in the space defined by identical vectors. Martin and Uriagereka (2014: 175) propose a similar perspective (some details about differences in implementation are discussed in Krivochen, 2015b):

> *Another analogy might help to understand the intuition we are entertaining. Imagine two sheets of origami paper, one black and the other grey, combined by stapling the bottom edge of the grey sheet to the top edge of the black sheet. We could continue this "derivation" by taking a new sheet of grey paper and stapling it to black sheet, so that now we have a total three sheets combined into a single object with the bottom-to-top order grey-black-grey* […]. *However, suppose that rather than introducing a new grey sheet, we instead fold the sheets from the initial step in such a way that we staple the bottom edge of the black sheet to the top edge of the grey sheet* […].

It is not obvious that phrase markers so conceived of are D&C computable, at least not in a way that does not prove to be insensitive to topological considerations which we have argued are essential in 'filler-gap' dependencies.

This brief discussion has attempted to highlight that there are plausible, theoretically attractive, and empirically useful alternatives to D&C approaches to structure building and structure mapping, which do not appeal to uniform monotonic phrase structure or sorting procedures to relate post-transformational to pre-transformational phrase markers[25]. We have argued that these alternatives are still necessary in the present stages of transformational generative grammar in order to derive semantic interpretation at different levels (binding and theta-theory being notable examples).

## 5. Conclusion:

In this paper we have argued that D&C algorithms are inadequate as models for natural language from two different perspectives:

[25] There are at least two other frameworks we have not mentioned here but which could potentially be added to the set of theories that are do not assume D&C mechanisms to 'infer' syntactic structure from a string: some versions of Combinatory Categorial Grammar (as pointed out to us by Shalom Lappin, p.c.) and Medeiros' (2015) Stack-Sorting grammar. Both, however, rely on what is essentially a Context-Free grammar and its corresponding PDA architecture, with a queue and a memory stack. It remains to be seen to what extent they are really radically different from LFG or HPSG, or even MGG (CCG, for instance, encodes notions like c-command and Binding principles by means of conditions over semantic representations understood in terms of Lambda calculus; see e.g. Steedman and Baldridge, 2011).

a) Structure building: empirical considerations show that different substrings of natural languages belong to different levels in the Chomsky Hierarchy. We have argued furthermore that natural languages display structural dynamicity; this dynamicity makes the Chomsky Hierarchy as it stands irrelevant for the study of natural languages. A single kind of sorting algorithm that attempts to assign a structural description to every string of a language falls short in some cases and overgenerates in others. Such an algorithm is procrustean in two respects, its failure to do justice to the mixed nature of linguistic structure being the first. The second respect is that it is insensitive to structural factors that are conditioned by semantics: we have seen such factors at work in scope ambiguities and variability of chunk size. These factors are intractable for a D&C algorithm.

b) Structure mapping: problems pertaining to structure mapping are somewhat more subtle than those pertaining to structure building; details vary depending on the version of transformational syntax one assumes. What seems beyond question is that no polystratal theory can avoid the reconstruction of pre-transformational levels in order to arrive at semantic interpretations. We have seen further that such reconstruction is not straightforward in phrase structure grammars; even if we assume a fully explicit Categorial Grammar hybrid, as in (46), further complications arise at the theory's core. Computational advantages that have been claimed for a lexical insertion + indexing model of grammar over a copy + deletion model (Berwick, 1984) are unclear: we have seen that an increase in the size of the grammar is required. Such increased complexity is unavoidable if the uniform approach to linguistic structure that is required for a D&C algorithm is assumed by axiom.

In general, and following the model proposed in Hopcroft and Ullman (1969: 211), the question whether given a grammar G, L(G) is D&C solvable (call it $Q_1$) has two characteristics *qua* computational question:

a) A question consists of an infinity of *instances* (i.e., there is an infinity of grammars, and each grammar generates an infinity of strings)
b) In each instance, the answer for the question is *yes* or *no*

What we have attempted to prove in this paper is that there are instances in natural languages for which the answer to $Q_1$ is clearly *no* on empirical grounds. Moreover, we would like to suggest that it is their intrinsic D&C algorithmic nature that makes Minimalism, Construction Grammar, Simpler Syntax, and HPSG mutually translatable (and subject to the same problems), as pointed out by Müller (2013).

In summary, the present paper has argued that D&C algorithms (and, more in general, a *function-based* approach to cognitive computation[26]) have a limited applicability for the modelling of natural languages, for they require the additional assumption of Structural Uniformity –which is empirically problematic- and the theories that assume such algorithms share limitations with respect not only to their empirical adequacy, but also to their theoretical modelling. The limitations we have focused on stem from two fundamental properties of natural language strings: their *lack of computational uniformity*—different substrings of a natural language occupy different positions on the

[26] For a wider perspective, which does not address specific properties of natural language but addresses the architecture of cognitive procedures in general, and also rejects the narrow equation 'effective computation = function-based computation' in favour of an interactive, dynamic interpretation of 'computation', see Goldin and Wegner (2007).

Chomsky Hierarchy—and natural languages' *structural dynamicity* in filler-gap dependencies. We have also examined difficulties faced by any sorting algorithm in the context of current theories of grammar, both transformational and non-transformational. Finally, we have indicated how these difficulties are avoided in a field-theoretic typological approach to syntactic structures as cognitive entities, which works as a possible alternative among others that, we argue, deserve to be explored.